\documentclass[10pt,twocolumn,letterpaper]{article}

\usepackage{iccv}
\usepackage{mmstyle}
\usepackage{times}
\usepackage{epsfig}
\usepackage{graphicx}
\usepackage{amsmath}
\usepackage{amssymb}
\usepackage{algorithm}
\usepackage{algorithmicx}
\usepackage{algpseudocode}
\usepackage{pifont}
\usepackage{comment}
\usepackage{mathtools}
\usepackage{bm}
\usepackage{bbm}

\usepackage{caption}
\usepackage{booktabs}  
\usepackage{threeparttable}  
\usepackage{multicol}  
\usepackage{multirow}  
\usepackage{tabu} 
\usepackage{bm}
\usepackage{float}
\usepackage[normalem]{ulem}
\usepackage[british,UKenglish,USenglish,english,american]{babel}

\newlength\savewidth\newcommand\shline{\noalign{\global\savewidth\arrayrulewidth
		\global\arrayrulewidth 1pt}\hline\noalign{\global\arrayrulewidth\savewidth}}

\newcommand{\Tref}[1]{Table~\ref{#1}}

\newcommand{\Fref}[1]{Figure~\ref{#1}}

\newcommand{\one}[1]{\mathbbm{1}_{[#1]}}

\usepackage[pagebackref=true,breaklinks=true,letterpaper=true,colorlinks,bookmarks=false]{hyperref}

\iccvfinalcopy 

\def\iccvPaperID{2954} 

\ificcvfinal\fi

\begin{document}
	
	\title{Improve Unsupervised Pretraining for Few-label Transfer}
	\author{Suichan Li$^{1,}$\thanks{Equal contribution, $\dagger$ Dongdong Chen is the corresponding author}, Dongdong Chen$^{2,*,\dagger}$, Yinpeng Chen$^{2}$, Lu Yuan$^{2}$, Lei Zhang$^{2}$, Qi Chu$^{1}$, Bin Liu$^{1}$, Nenghai Yu$^{1}$, \\
		$^{1}$University of Science and Technology of China 
		\quad\quad $^{2}$Microsoft Research\\
		\{lsc1230@mail., qchu@, flowice@, ynh@\}ustc.edu.cn, cddlyf@gmail.com, \\
		\{ yiche, luyuan, leizhang\}@microsoft.com}
	\maketitle
	\ificcvfinal\thispagestyle{empty}\fi
	
	\begin{abstract}
		Unsupervised pretraining has achieved great success and many recent works have shown unsupervised pretraining can achieve comparable or even slightly better transfer performance than supervised pretraining on downstream target datasets. But in this paper, we find this conclusion may not hold when the target dataset has very few labeled samples for finetuning, \ie, few-label transfer. We analyze the possible reason from the clustering perspective: 1) The clustering quality of target samples is of great importance to few-label transfer; 2) Though contrastive learning is essential to learn how to cluster, its clustering quality is still inferior to supervised pretraining due to lack of label supervision.  Based on the analysis, we interestingly discover that only involving some unlabeled target domain into the unsupervised pretraining can improve the clustering quality, subsequently reducing the transfer performance gap with supervised pretraining. This finding also motivates us to propose a new progressive few-label transfer algorithm for real applications, which aims to maximize the transfer performance under a limited annotation budget. To support our analysis and proposed method, we conduct extensive experiments on nine different target datasets. Experimental results show our proposed method can significantly boost the few-label transfer performance of unsupervised pretraining.
		
	\end{abstract}

	\vspace{-1em}
	\section{Introduction}
	\begin{figure}[t]
		\centering
		\includegraphics[width=0.9\linewidth]{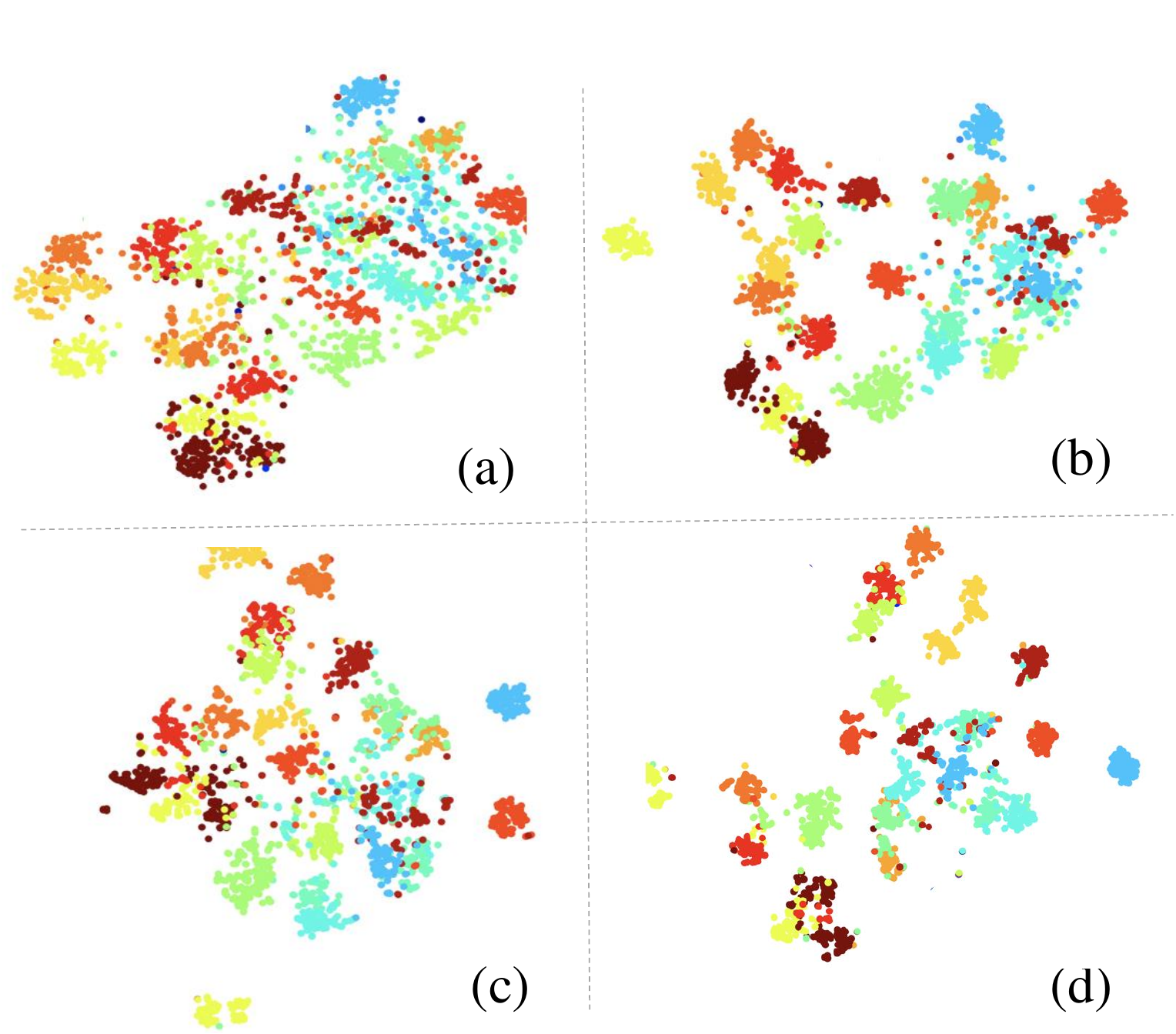}
		\caption{\small t-SNE visualization of features on Pet~\cite{parkhi2012cats} by using different models: (a) unsupervised pretrained model, (b)supervised pretrained model, (c)  target-aware unsupervised pretrained(TUP) model, (d) finetuned TUP model by using a few labeled samples. } \label{fig:case}
		\vspace{-2em}
	\end{figure}
	
	Model pretraining plays a key role for deep transfer learning. By pretraining the model on a large auxiliary source dataset and then fine-tuning on the small-scale target dataset, it can achieve better performance than the train-from-scratch counterpart. The recent work BiT~\cite{kolesnikov2019big} has shown that supervised pretraining on large scale source dataset can achieve very strong transfer performance. Despite the great success of supervised pretraining, a large amount of labeled source data is required. Recently, unsupervised pretraining \cite{he2020momentum,chen2020simple,grill2020bootstrap,caron2020unsupervised,chen2020big,byol_2020} has achieved great progress. By directly pretraining on the larger-scale unlabeled data (\eg, ImageNet), many state-of-the-art (SOTA) unsupervised learning works \cite{chen2020simple,grill2020bootstrap,byol_2020,caron2020unsupervised} demonstrate that unsupervised pretraining can achieve comparable or even slightly better transfer performance than supervised pretraining on many downstream target datasets. 
	
	In this paper, we ask the question ``does unsupervised pretraining really achieves comparable transfer performance as supervised pretraining?". And we empirically find the answer is ``no'' when the downstream target dataset has limited label samples for finetuning, \ie, \emph{``few-label transfer''}. We seek to investigate the underlying reason from the clustering perspective. We hypothesize that the clustering of target samples in the feature space is of great importance for few-label transfer and unsupervised pretraining has worse clustering quality than supervised pretraining. Intuitively, if the pretrained representation has a very good clustering in the target space, it will only need very few labels to learn a good classifier boundary. To verify our hypothesis, we compare the clustering quality of unsupervised and supervised pretrained models on the target dataset in \Fref{fig:case} (a) (b). Obviously, the target samples are better clustered by using the supervised pretrained models. The following analysis (\Tref{tab:cluster_abl}) will also show the positive correlation between the clustering quality and the few-label transfer performance.
	
	To understand why unsupervised pretraining has inferior clustering quality, we follow the work \cite{wang2020understanding} to analyze the widely used contrastive loss. Specifically, the contrastive loss can be decomposed into two terms: an \emph{alignment} term that encourages two samples of a positive pair should be as close as possible, and a \emph{uniformity} term that encourages the learned representation to uniformly distribute on the unit hypersphere. With the alignment term, by using strong augmentation during training, the sub-space of similar images will overlap and be pulled closer. In other words, contrastive learning is trying to cluster the pretraining unlabeled data, but it encourages the learned representation to distribute in the whole space. Therefore, if the target data has a large domain gap with the source data, their feature representations will scatter in the whole space and hard to cluster. By contrast, supervised pretraining does not encourage the learned representation to be uniformly distributed and the label supervision also provides stronger alignment force across different images. So the learned representation is more compact and better clustered even for the same target domain.
	
	Based on the above analysis, we discover that only involving some unlabeled target data into the unsupervised pretraining process (\emph{``target-aware''unsupervised pretraining, or \textbf{TUP}}) can significantly improve its clustering quality (\Fref{fig:case} (c)), thus subsequently reducing the performance gap with supervised pretraining. This finding is very interesting and useful in real application scenarios where some small-scale unlabeled data is easy to obtain. On the other hand, considering data annotation is often conducted after unlabeled data collection, we further study the question that \emph{``can we leverage the clustering property to maximize the target performance under a limited annotation budget''}. And we propose a simple progressive few-label transfer algorithm for practical usage. Specifically, given the pretrained representation, we first conduct the clustering on the unlabeled target data to find the most representative samples to annotate, and then use the annotated samples to finetune the pretrained model. The finetuned model can further improve the clustering quality (\Fref{fig:case} (d)), thus making data annotation and model finetuning form an active co-evolution loop.
	
	To demonstrate our finding and the proposed method, extensive experiments are conducted on nine different target datasets. The experimental results demonstrate that the proposed method can significantly improve the few-label transfer performance for unsupervised pretraining, and even outperform supervised pretraining. For example, when each target dataset has 10 labeled samples per category, our proposed TUP can boost the average transfer performance of unsupervised pretraining from 67.49\% to 74.15\%, slightly better than supervised pretraining 73.27\%. By further equipping our progressive transfer strategy, the transfer performance can increase to 76.69\% under the same annotation budget. To summarize, our contributions are three-fold: 1) We are the first that points out the few-label transfer gap between unsupervised pretraining and supervised pretraining, which is not studied in the research field yet; 2) We analyzed the possible underlying reasons and discover a simple and effective strategy for real applications where some small-scale unlabeled data can be collected; 3) We further propose a progressive few-label transfer strategy to boost the performance under the limited annotation budget.
	
	\section{Related works}
	\paragraph{Supervised Pretraining and Unsupervised Pretraining.} Model pretraining is very important in the deep learning literature. Before the surgence of unsupervised pretraining, the main success and study focused on supervised pretraining \cite{he2016deep,he2019rethinking}. And the work BiT~\cite{kolesnikov2019big} shown large scale supervised pretraining is very effective on downstream tasks. In recent two years, the representative works \cite{wu2018unsupervised,he2020momentum,chen2020simple} ignited the interests of the 
	research field in studying unsupervised pretraining, and made great progress \cite{caron2020unsupervised,byol_2020,chen2020big,fu2021unsupervised}. By evaluating the performance on many downstream target datasets, they demonstrate that unsupervised pretraining has shown comparable transfer performance to supervised pretraining. In this paper, we find this conclusion does not hold when the target dataset has few labeled samples for finetuning. Our work is complementary to existing unsupervised pretraining works, and proposed two practical strategies to improve the transfer performance for real applications.
	
	\vspace{0.5em}
	\noindent\textbf{Few-shot Learning and Active Learning.} Though our focus is to analyze the transfer performance of unsupervised pretraining, our work is loosely related to few-shot learning ~\cite{FinnAL17,Sachin2017,mishra2018a} and may benefit the finetuning based few-shot learning methods ~\cite{chen2018a,Dhillon2020A}. We demonstrate that, if some small-scale unlabeled data exists in the target domain, we can leverage it to improve the pretrained representation and can achieve better few-shot performance. For semi-supervised learning  \cite{li2020density,sohn2020fixmatch,berthelot2019remixmatch}, the improved pretrained representation can also provide better initialization and 
	boost the performance. Our progressive transfer strategy shares the similar spirit as the classical active learning \cite{lewis1995sequential,roy2001toward,bilgic2009link,beluch2018power,gal2017deep}. However, most active learning methods only consider the target domain and involve very complicated sampling strategies. In this paper, we aim to improve the transfer performance from both the pretraining and transfer perspective and propose a simple and effective strategy. But we believe combining our method with more sophisticated active learning strategies can achieve better performance, which we leave for future study. 
	\begin{table}[t]
		\small
		\begin{center}
			\setlength{\tabcolsep}{0.3mm}{
				\begin{tabular}{c|l|cccccc}  
					& Method 
					& DTD   & Food101  & CIFAR10  & CIFAR100  & EuroSAT & Pet37 \cr 
					\shline 
					\multirow{4}{*}{2}& MoCoV2 
					& 38.06 & 15.80   &  41.29   & 22.95     & 60.20  & 56.87  \cr
					& DCV2
					& 37.18     & 20.08       & 43.12        & 22.36         & 53.84      & 49.38  \cr
					& SimCLRV2
					& 35.23    & 17.44        & 40.52        & 18.33         & 57.17      & 37.40  \cr 
					& BiT 
					& \textbf{44.66} & \textbf{24.99}    & \textbf{59.05}    & \textbf{37.40}     & \textbf{68.31}  & \textbf{63.95}  \cr 
					\hline
					\multirow{4}{*}{4} & MoCoV2  
					& 48.20  & 25.49   & 50.30    & 35.08     & 69.18 & 68.94  \cr
					& DCV2
					& 46.47    & 30.63       & 49.90        & 33.11         & 58.34      & 57.28  \cr
					& SimCLRV2
					& 46.54    & 27.42        & 48.78        & 28.80         & 70.42      & 51.12  \cr 
					& BiT 
					& \textbf{53.69}  & \textbf{35.26}   & \textbf{71.90}    & \textbf{47.79}  & \textbf{79.05} & \textbf{76.08} \cr
					\hline
					\multirow{4}{*}{6} & MoCoV2 
					& 53.00   & 31.01    & 56.44    & 42.74    & 72.15  & 72.35  \cr
					& DCV2
					& 50.93     & 36.51       & 49.57        & 39.70         & 65.00      & 60.09  \cr
					& SimCLRV2
					& 50.90    & 34.08        & 53.59        & 35.42         & 72.97      & 58.87  \cr 
					& BiT 
					& \textbf{57.64}  & \textbf{41.26}     & \textbf{75.62}    & \textbf{54.17}  & \textbf{82.55}   & \textbf{79.85} \cr 
					\hline
					\multirow{4}{*}{10} & MoCoV2 
					& 58.68  & 38.38  & 59.92    & 51.87    & 75.69  & 77.62  \cr
					& DCV2
					& 56.32     & 43.92       & 53.06        & 48.06         & 75.26      & 64.24  \cr
					& SimCLRV2
					& 55.89    & 42.10        & 59.64        & 43.57         & 75.77      & 69.11  \cr 
					& BiT 
					& \textbf{63.06}  & \textbf{48.77}  & \textbf{79.99}    & \textbf{59.92}    & \textbf{86.25}  & \textbf{84.81}   \cr 
			\end{tabular} }
			\caption{The few-label transfer performance on six different target datasets for three SOTA unsupervised pretrained models, including MoCoV2~\cite{chen2020improved}, SimCLRv2~\cite{chen2020big} and DeepClusterV2 (DCV2) \cite{caron2020unsupervised}, and supervised pretrained models from BiT ~\cite{kolesnikov2019big}. All the results are averaged by 5 trials. }
			\label{tab:transfer_study}
		\end{center}
		\vspace{-3em}
	\end{table}
	\section{Few-Transfer Analysis}
	We formulate the problem of few-label transfer in the paradigm of pretraining and finetuning. Under the supervised pretraining setting, the model is first pretrained on a large-scale labeled source dataset $\mathcal{S}^\#=\{x^s_i, y^s_i\}_{i=1}^M$, and then finetune the model on the small-scale target dataset $\mathcal{T}=\{x_j, y_j\}^N_{j=1}$ with few labeled samples, where $N\ll M$. Under the unsupervised pretraining setting, the source dataset $\mathcal{S}=\{x^s_i\}_{i=1}^M$ is fully unlabeled and the target dataset $\mathcal{T}$ is the same.
	
	To compare the few-label transfer ability we adopt three existing SOTA unsupervised pretraining methods, \ie, MoCoV2~\cite{chen2020improved}, SimCLRV2\cite{chen2020big} and DeepClusterV2(DCV2)\cite{caron2020unsupervised}, and the supervised pretraining models from BiT~\cite{kolesnikov2019big}. The ImageNet~\cite{deng2009imagenet} is used as the large-scale source data for pretraining, and the subsets of six small-scale target datasets are used for few-label transfer: Pet37~\cite{parkhi2012cats}, DTD~\cite{cimpoi2014describing}, CIFAR10 and CIFAR100~\cite{krizhevsky2009learning}, Food101~\cite{bossard2014food} and EuroSAT~\cite{helber2019eurosat}. The detailed comparison results are shown in the Tab.\ref{tab:transfer_study}. It can be seen that all the unsupervised pretrained models show inferior few-label transfer performance than the supervised counterpart.
	
	\begin{figure}[t]
		\centering
		\includegraphics[width=1.0\linewidth]{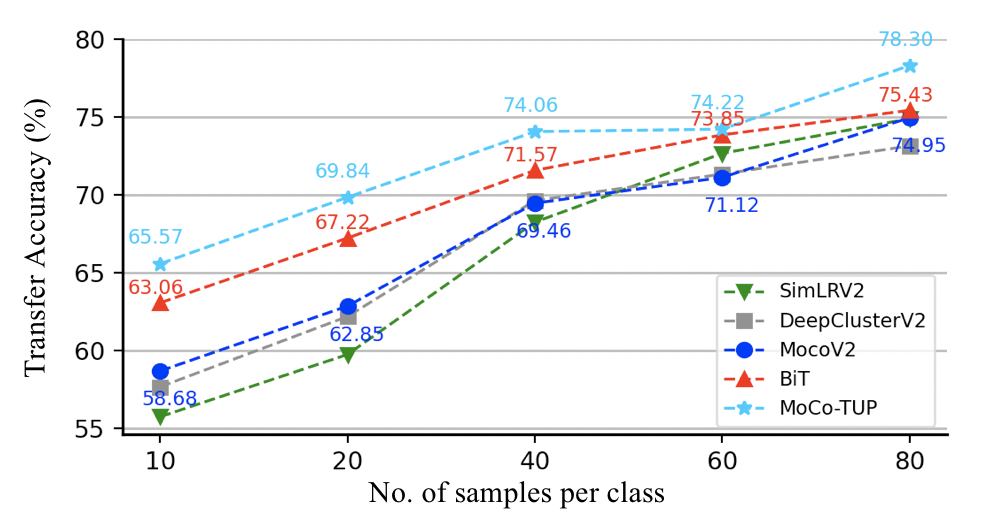}
		\caption{The transfer performance comparison by varying different number of labeled samples during finetuning, ``MoCo-TUP'' is our target-aware unsupervised pretraining.}
		\label{fig:label_vary_2}
	\end{figure}
	
	We continue to compare the transfer performance between unsupervised pretraining and supervised pretraining in depth, from few-label transfer to full-label transfer. Here, full-label transfer means finetuning the pretrained model on the full labeled target dataset. Specifically, we take the DTD dataset (80 labeled samples per category) as an example, and test the transfer performance of different pretrained models by varying the number of labeled samples during finetuning. As shown in Figure \ref{fig:label_vary_2}, although supervised pretraining is superior to the unsupervised pretraining by a considerable margin, the performance gap becomes much smaller as the number of labeled samples increases. A similar trend can also be observed in other datasets. Therefore, we have the following observations:
	\begin{itemize}
		\item The unsupervised pretrained representation itself is not bad. Given a moderate number of labels, it can match or even beat the transfer performance of the supervised counterpart. This is consistent with the full-label transfer conclusion in existing unsupervised learning works~\cite{chen2020simple,grill2020bootstrap}. 
		\item But for few-label transfer,  unsupervised pretraining is often inferior to supervised pretraining.
	\end{itemize}
	
	\begin{table}[t]
		\small
		\begin{center}
			\setlength{\tabcolsep}{1.0mm}{
				\begin{tabular}{llccc|c}  
					Dataset  &   & Unsup-1k & Sup-100 & Sup-1k & TUP  \\
					\shline
					\multirow{2}{*}{Pet} & Cluster Acc. &    47.72      &    12.82     &  67.44      &  61.69 \\
					& Transfer Acc. & 70.93 & 45.65 & 77.94 & 75.10 \\
					\hline
					\multirow{2}{*}{Food101} & Cluster Acc & 11.86      &    4.03     &  17.23      &  39.23 \\
					& Transfer Acc. & 28.39 & 18.29 & 38.69 & 52.25 \\
			\end{tabular} }
			\captionof{table}{The clustering accuracy and few-label transfer performance of different pretrained models. ``Unsup, Sup, TUP" are unsupervised, supervised and target-aware unsupervised pretraining respectively, and ``-1k/100" is ImageNet-1k/100.} 
			\label{tab:cluster_abl}
		\end{center}
		\vspace{-3em}
	\end{table}
	
	\noindent \textbf{Clustering Matters to Few-label Transfer.} We propose a clustering perspective to analyse the reason why unsupervised pretraining shows poor few-label transfer performance than supervised pretraining. Here we use MoCoV2\cite{chen2020improved} as the instantiation of unsupervised pretraining,  and compare unsupervised and supervised pretraining in terms of the target sample distribution in the feature space through t-SNE \cite{hinton2002stochastic}. The visualization is shown in~\Fref{fig:case} and the target Pet37 dataset~\cite{parkhi2012cats} is used as the example. As we can see in~\Fref{fig:case} (a) and (b), the features obtained from the supervised pretrained model are better clustered than those obtained from the unsupervised pretrained model. Based on this observation, we make intuitive sense that \textit{the clustering quality matters to few-label transfer}.
	
	This hypothesis can be further elucidated. If the target features are well clustered after pretraining, it is much easier to learn a good classifier even though only a few labeled samples are available in the following finetuning. We further quantitatively study the relationship between the clustering quality and few-label transfer performance with only 5 labeled samples per class on different pretrained models: unsupervised pretrained ResNet-50 on ImageNet-1k, supervised pretrained ResNet-50 on ImageNet-100(a subset of ImageNet-1K with 100 categories) and ImageNet-1k respectively. We use the BCubed Precision (Cluster Acc) as the metric of clustering quality~\cite{amigo2009comparison}. The results shown in \Tref{tab:cluster_abl} demonstrate that the few-label transfer performance has a positive correlation to the clustering accuracy. 
	
	\begin{figure*}[!t]
		\centering
		\includegraphics[width=1.0\linewidth]{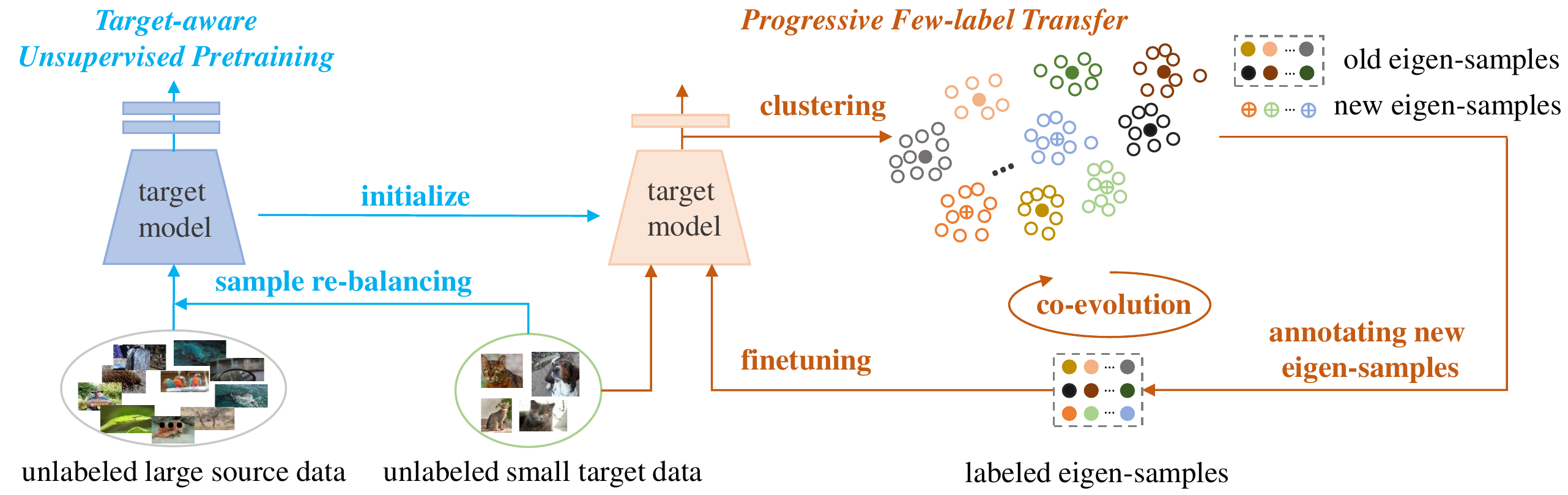}
		\caption{The improved unsupervised pretraining framework for few-label transfer in real applications, which has two key components: target-aware unsupervised pretraining and progressive few-label transfer. By involving target data into pretraining, target-aware unsupervised pretraining can get better clustering in the target space. Progressive few-label transfer co-evolves the process of eigen-sample selecting by clustering and model-finetuning.} 
		\label{fig:pipeline}
		\vspace{-1em}
	\end{figure*}
	
	\vspace{0.5em}
	\noindent\textbf{Understand Contrastive Learning.} To further understand why unsupervised pretraining has worse clustering quality than supervised pretraining, we follow \cite{wang2020understanding,chen2020intriguing} and decouple the widely used unsupervised learning loss, \ie, contrastive loss, into two terms: one \emph{alignment} term and one \emph{uniformity} term. Formally, following the definition in \cite{chen2020simple}, the contrastive loss between two augmentations $(i,j)$ of the same image for a mini-batch $\mathcal{B}$ is: 
	\begin{equation}
	\label{eqn:contrastive}
	\mathcal{L}_{ctr} = -\frac{1}{N}\sum_{i,j\in \mathcal{B}}\log \frac{\exp(\mathrm{sim}(\bm z_i, \bm z_j)/\tau)}{\sum_{k=1}^{2N} \one{k \neq i}\exp(\mathrm{sim}(\bm z_i, \bm z_k)/\tau)}
	\end{equation}
	where $\bm z_i, \bm z_j$ are the normalized representations extracted from the target model for the two augmented views of the same example. $\mathrm{sim}(\bm u, \bm v)$ is the cosine similarity between $\bm u$ and $\bm v$. $N$ is the batch size and $\tau$ is the temperature hyper-parameter. By expanding the loss, the above loss can be rewritten as:
	\begin{equation}
	\begin{aligned}
	\label{eqn:split_contastive}
	\mathcal{L}_{ctr} &= -\frac{1}{N\tau}\sum_{i,j}\mathrm{sim}(\bm z_i, \bm z_j)
	+ \\&\frac{1}{N}\sum_i\log\sum_{k=1}^{2N} \one{k \neq i}\exp(\mathrm{sim}(\bm z_i, \bm z_k)/\tau)
	\end{aligned}
	\end{equation}
	The first term of Eq.\ref{eqn:split_contastive} is the \textit{alignment} term, which encourages two augmentations of each image in the mini-batch (a positive pair) to have similar features. By using strong augmentation during training, the sub-space of similar images will overlap and be pulled closer. The second term is closely connected to the pairwise potential in a Gaussian kernel and can be minimized with a perfect uniform encoder, thus named as the \textit{uniformity} term. The uniformity term encourages the feature vectors to be roughly uniformly distributed on the unit hypersphere (the normalized whole feature space). In this sense, we can find that contrastive learning is indeed to cluster the pretraining unlabeled data, but it encourages the learned representation to uniformly distribute in the whole space. Therefore, if the target dataset has some domain gap with the source dataset, their feature representations will scatter and hard to cluster. By contrast, there is no such uniformity term in the supervised pretraining and the label supervision can also provide stronger alignment force across different images than the alignment force from two augmentations of the same image in the contrastive loss. Therefore, the supervised pretrained representation may reside in a more compact space.
	
	\section{Target-aware Unsupervised Pretraining}
	\noindent\textbf{Target-aware Unsupervised Pretraining.} Based on the above analysis, in order to boost the few-label transfer performance of unsupervised pretraining, we should improve the clustering quality of the pretrained representation in the target domain. Considering contrastive learning is able to cluster the pretraining unlabeled data, we propose a simple and effective strategy called \emph{Target-aware Unsupervised Pretraining}(TUP). It is designed for the typical applications where some small-scale unlabeled data is relatively easy to obtain. Specifically, besides the large-scale source data, we also add the unlabeled target data into the unsupervised pretraining stage, so that the unsupervised pretrained model can also have a better clustering quality of the target data. By contrast, existing unsupervised pretraining that only utilizes the source data can be regarded as ``target-agnostic''. The improved clustering will significantly boost the transfer performance. In \Tref{tab:cluster_abl}, we show TUP's clustering accuracy and its corresponding transfer performance on the target domain, and the corresponding feature visualization is shown in \Fref{fig:case} (c).
	
	\vspace{0.5em}
	\noindent\textbf{Sample Re-balancing.} Empirically, we find naively mixing the small-scale unlabeled target data  and the large-scale unlabeled source data with the ratio $1:1$ in the pretraining does not work well. Because the amount of the unlabeled target images in $\mathcal{T}$ is much smaller than the auxiliary source dataset $\mathcal{S}$, it will cause serious learning imbalance and make target-aware unsupervised pretraining degrade to the vanilla unsupervised pretraining. To mitigate this issue, we propose a simple and effective sample re-balancing strategy which increases the percentage $p$ of target data in the mixture of target data $\mathcal{T}$ and source data $\mathcal{S}$. Besides, we observe finding a proper percentage $p$ is necessary, a too large or small percentage $p$ will both cause the degradation of performance, which we will study in the ablation part.
	
	\begin{algorithm}[t]
		\small
		\caption{\small{Anchor-Constrained KMeans in $\kappa$-th evolution}}
		\renewcommand{\algorithmicrequire}{\textbf{Input:}}
		\renewcommand{\algorithmicensure}{\textbf{Output:}}
		\begin{algorithmic}[1]
			\Require Set of target features $\cF=\{f_i\}^N_{i=1}$ (estimated cluster label of $f_i$ is denoted as $f_i^{L}$). Number of new clusters $K$. Set of anchors $\cA=\{a_j\}^m_{j=1}$, Maximum iteration of KMeans $t_{max}$.
			\Ensure $K$ cluster centers $\{\mu_j\}_{j=1}^{m+K}$
			\State \textbf{-- Initialize Centers:}
			\State ${\mu_{j}^0}\xleftarrow[]{}{a_j}$, $j = 1, ..., m$; randomly initialize $\mu_{m+1}^0,...,\mu_{m+K}^0$.
			\For{$ t = 1,\dots, t_{max} $}
			\State \textbf{-- Assign Samples to Cluster}: 
			\For{$ i = 1,\dots, N $} 
			\State $f_i^{L}=\text{arg} ~ \underset{j}{\text{min}} ~ ||f_i-\mu_{j}||^2$, $j=1,...,m+K$
			\EndFor
			\State \textbf{-- Update Cluster Centers:}
			\State $\mu_j^{t}=\mu_j^{t-1}$, $j=1,...,m$
			\For{$ j = m+1,\dots, m+K $}
			\State $\cF_j^t=\{f_i|f_i^{L}=j, f_i\in \cF\}$,\vspace{0.3em}
			\State $\mu_j^{t}=\frac{1}{|\cF_j^t|}\sum_{f\in\cF_j^t}{f}$,
			\EndFor
			\EndFor
		\end{algorithmic}
		\label{alg:AC_KMeans}
	\end{algorithm}
	
	\section{Progressive Few-label Transfer}
	Since data annotation is often conducted after the unlabeled data collection in real applications, the relationship between the few-label transfer performance and the clustering quality further motivates us to study the question ``can we leverage this property to maximize the target performance under a limited annotation budget''. This is important for the applications where data annotation is extremely difficult and costly. Before elaborating our final strategy, we first introduce our motivations from two perspectives:
	\begin{itemize}
		\item The target samples closer to the clustering centers are more representative (called ``eigen-samples''), which suggests choosing such samples to label can be more effective, especially under a very limited label budget.
		\item Finetuning the model with such labeled samples can further improve the clustering quality of all target samples, and in return, the improved model continues to help identify more representative samples.
	\end{itemize}

	\noindent\textbf{Progressive Few-label Transfer.} Integrated with the above motivations, we propose a new \emph{progressive few-label transfer} strategy for real applications. As shown in \Fref{fig:pipeline}, the progressive few-label transfer follows a co-evolution process: {``\textit{clustering} $\rightarrow$  \textit{eigen-samples annotation} $\rightarrow$ \textit{model finetuning} in a loopy way"}. Specifically, at each evolution step $\kappa$, we first re-cluster the target features and incrementally find some eigen-samples, then annotate the new eigen-samples, and finally finetune the model with all the labeled eigen-samples. This co-evolution process will end until we reach the total annotation budget. 
	
	We develop a new KMeans-based clustering algorithm called \textbf{A}nchor \textbf{C}onstrained \textbf{KMeans} (\textbf{ACKMeans}) to implement the incremental eigen sampling. All eigen-samples found at previous $\kappa-1$ evolution steps are referred to \textit{anchors}. The key idea of ACKMeans (at $\kappa$-th evolution) is that the anchors as cluster centers won't be changed during KMeans and help exclude samples close to these anchors; while the remaining of dissimilar samples would be clustered into $K$ new clusters, which helps select $K$ new eigen-samples to annotate. This way allows us to optimize the annotation budget to the most extent, since each eigen-sample represents a cluster of similar samples associated to it. At every evolution, supposing $b$ annotation budget per category, totally $K=b \times C$ new eigen-samples are chosen to be annotated, where $C$ denotes the number of target categories. Hence, the total annotation budget for target data would be $K \times \kappa_{max}$, where $\kappa_{max}$ is the maximum evolution steps. The Alg.\ref{alg:AC_KMeans} shows the detailed procedure of ACKMeans.
	
	To apply the progressive few-label transfer strategy in the real applications, we suggest a practical ``$1+\epsilon$'' setting. Initially, only ``$1$" image per target category is given. We think it reasonable since each category needs an indicator image when the annotation process starts. Next, ``$\epsilon$" extra annotations are required to be labeled for each category on average, and thus the total annotation cost $\epsilon \times C = K \times \kappa_{max}$. In this setting, we do not guarantee each category can get exactly $\epsilon$ extra labels. 
	
	By contrast, existing few-label transfer setting assumes all the labels to be pre-known or randomly chooses a certain percentage (\eg, $5\%$) of labeled samples per category to guarantee number of labeled samples are class balanced. It can be regarded as an \textbf{``oracle setting''}, since consuming labels beforehand is usually unrealistic in real applications or it will need the annotator to watch and label more samples beyond the few labels. 
	\begin{table*}[!t]
		\small
		\begin{center}
			\setlength{\tabcolsep}{1.9mm}{
				\begin{tabular}{l|l|ccccccccc|c}  
					1+ $\epsilon$ & Method 
					& DTD   & Food101 & SUN397    & Caltech101 & STL10  & CIFAR10  & CIFAR100  & EuroSAT & Pet37 & Mean Acc.\cr 
					\shline 
					\multirow{4}{*}{1+1}& MoCoV2 
					& 38.06 & 15.80    & 24.28   & 63.12      & 75.06   &  41.29   & 22.95     & 60.20  & 56.87 & 45.03 \cr
					& BiT 
					& 44.66 & 24.99   & 27.21    & 61.07      & 74.80   & 59.05    & \textbf{37.40}     & 68.31  & 63.95 &  52.70 \cr 
					& Ours
					&   44.79   & 34.27   & 33.16   & 78.75   & 81.74   & 59.44    & 30.24     & 68.84  & 65.93  & 55.24 \cr 
					& Ours-Pro  
					& \textbf{44.86} & \textbf{35.92}   & \textbf{35.46}   & \textbf{79.87}   & \textbf{82.45}   & \textbf{62.19}    & 32.74     & \textbf{69.62}  & \textbf{68.87}  & \textbf{59.30} \cr 
					\hline
					\multirow{4}{*}{1+3} & MoCoV2  
					& 48.20  & 25.49  & 35.06   & 76.54       & 87.35    & 50.30    & 35.08     & 69.18 & 68.94 &  56.84 \cr
					& BiT 
					& 53.69  & 35.26  & 36.72   & 73.89      & 83.28     & 71.90    & \textbf{47.79}  & \textbf{79.05} & \textbf{76.08} &  63.61\cr 
					& Ours 
					&   55.17  & 48.03  & 43.30   & 84.92  & 88.07     & \textbf{72.86}    & 43.94    & 76.15   & 73.65  & 65.12 \cr
					& Ours-Pro 
					& \textbf{57.11}  & \textbf{49.66}  & \textbf{45.38}   & \textbf{86.13}  & \textbf{88.19}     & 71.20    & 46.80    & 76.70   & 74.07 & \textbf{68.51} \cr
					\hline
					\multirow{4}{*}{1+5} & MoCoV2 
					& 53.00   & 31.01  & 40.71   & 82.14      & 89.21     & 56.44    & 42.74    & 72.15  & 72.35 & 62.05 \cr
					& BiT 
					& 57.64  & 41.26  & 41.11    & 80.41      & 86.71     & 75.62    & \textbf{54.17}  & \textbf{82.55}   & \textbf{79.85} &  68.30 \cr
					& Ours
					&    60.05   & 55.35  & 47.85  & 87.23   & 90.10     & \textbf{77.22}    & 52.68   & 80.58   & 76.18 & 70.22   \cr 
					& Ours-Pro  
					& \textbf{61.56}  & \textbf{56.50}  & \textbf{49.42}  & \textbf{88.11}   & \textbf{90.14}     & 74.89    & 53.74   & 79.29   & 79.01 & \textbf{72.55}   \cr 
					\hline
					\multirow{4}{*}{1+9} & MoCoV2 
					& 58.68  & 38.38  & 47.75   & 86.75      & 90.85     & 59.92    & 51.87    & 75.69  & 77.62 &  67.49 \cr
					& BiT 
					& 63.06  & 48.77  & 44.96   & 86.29      & 90.07     & 79.99    & 59.92    & \textbf{86.25}  & \textbf{84.81}  & 73.27 \cr  
					& Ours
					& 65.57  & 62.56  & 53.48   & 89.19      & 91.39     & 79.67    & \textbf{60.91}    & 84.14   & 80.48  & 74.15 \cr 
					& Ours-Pro  
					& \textbf{66.58}  & \textbf{62.67}  & \textbf{54.30}   & \textbf{89.55}      & \textbf{92.02}     & \textbf{80.73}    & 60.63   & 81.42   & 84.41  & \textbf{76.69} \cr 
			\end{tabular} }
			\captionof{table}{\small  \textbf{The few-label transfer results on nine benchmark target datasets.} ``Ours-Pro'' means using our progressive few-label transfer strategy and ``Ours'' means using the oracle few-label transfer setting. All the results are averaged by 5 trials to reduce randomness.} 
			\label{tab:overall_results}
		\end{center}
		\vspace{-2em}
	\end{table*}
	
	\section{Experiments}
	\subsection{Experimental Setup}
	\noindent \textbf{Datasets.} In the following experiments, we use the ImageNet-1k dataset~\cite{deng2009imagenet} as the auxiliary large scale source dataset, and consider $9$ small-scale target datasets: Pet37~\cite{parkhi2012cats}, SUN397~\cite{xiao2010sun}, DTD~\cite{cimpoi2014describing}, CIFAR10 and CIFAR100~\cite{krizhevsky2009learning}, Caltech101~\cite{fei2004learning}, Food101~\cite{bossard2014food} and EuroSAT~\cite{helber2019eurosat}. These datasets are very diverse and differ in the total image number, input resolution and nature of their categories, ranging from general object categories (\eg, CIFAR10/100) to fine-grained
	ones (\eg, Pet37). We follow the standard setting as \cite{chen2020simple,grill2020bootstrap,kolesnikov2019big}, and report the mean class accuracy for Pet37, Caltech101 and the Top1 accuracy for other datasets. All the results are averaged by 5 trials to reduce randomness.
	
	\vspace{0.25em}
	\noindent \textbf{Pretraining Details.} We build our target-aware unsupervised learning based on MoCoV2~\cite{chen2020improved} and follow its training protocol. In details, we adopt the SGD optimizer with momentum $0.9$ and the weight decay $0.0001$. The initial learning rate is $0.24$ with a cosine scheduler and the batch size is 2,048. All the pretraining models are trained with 800 epochs. The backbone network for all the experiments uses ResNet-50~\cite{he2016deep}. The default sample re-balancing ratio varies based on the target dataset size so that the resampled target data size is about $20\%$ of the source dataset size.
	
	\vspace{0.25em}
	\noindent \textbf{Finetuning Details.} We finetune the pretrained model for $60$ epochs without weight decay. The learning rate for the newly added \textit{FC} layer and pretrained layers is $3.0$ and $0.0001$ respectively. We only use random crop with resizing, flips for training and the center crop with resizing for testing.
	
	\subsection{Overall Results}
	\Tref{tab:overall_results} reports the few-label transfer performance of the proposed model on all benchmark datasets. For comparison, we consider two models as our strong baseline: vanilla unsupervised pretrained models (MoCoV2~\cite{chen2020improved}) and the supervised pretrained models (BiT~\cite{kolesnikov2019big}) under the oracle labeling setting, which select ``$1+\epsilon$'' labeled samples for each category in a strictly class-balance way. We report both the performance under the oracle setting and the progressive ``$1+\epsilon$'' few-label transfer setting for our proposed method. Here, our method adopts the exactly same pretraining and finetuning setting to MoCoV2, and directly uses the officially released code for BiT pretraining and finetuning.
	
	We can observe the following main results. 1) Our target-aware unsupervised pretraining consistently outperforms the vanilla unsupervised pretraining baseline MoCoV2 across all the datasets by a large margin. The results verify the effectiveness of involving target set with source set into the unsupervised pretraining. 2) Our method outperforms the supervised pretraining(BiT) on majority of datasets and is comparable or slightly worse on the rest. On average, our method performs better than BiT. This also shows that a large amount of labeling information is very useful, and that target-aware pretraining can compensate for the gap caused by the lack of labeling information. 3) Combining target-aware unsupervised pretraining with the progressive few-label transfer can achieve better performance than the counterpart under the oracle setting, even though our practical ``1+$\epsilon$'' setting does not assume the class-balance.
	
	By analyzing the performance among different datasets, we further get some fine-grained observations:
	
	\textit{1) Our method outperforms both vanilla unsupervised pretraining and supervised pretraining when the gap between source and target domains is either very large (\eg, SUN397) or very small (\eg. Caltech101)}.
	For example, SUN397 is for scene recognition while ImageNet is almost object-centric. Therefore, either the supervised pretraining model or the vanilla unsupervised pretraining model cannot obtain good clustering on the target domain. (their Cluster ACC~\cite{amigo2009comparison}: $22.93\%$ \textit{vs.} $20.11\%$). In contrast, Caltech101 is object-centeric and shares similar categories with ImageNet, therefore both the supervised and the vanilla unsupervised pretrainings on ImageNet can achieve good clustering (their Cluster ACC: $47.11\%$ \textit{vs.} $53.14\%$). By involving the target data, our method can improve the clustering quality (Cluster ACC: $34.36\%$ on SUN397, $59.88\%$ on Caltech101) especially for large domain gap (SUN397), thus bringing significant performance gain. 
	
	\textit{2) Though our method only requires a small-scale unlabeled target dataset, we empirically find it will bring more benefits if the target dataset has a larger scale.} One typical example is the Food101 dataset. It has a total of about 75k high-resolution images and each category has about 750 images. It is consistent with the common sense that bigger data can help learn better representation.
	
	\textit{3) Our method is comparable to or slightly worse than supervised pretraining if the target dataset has a low image resolution.} For example, the image resolution of CIFAR10/100 and EuroSAT is only $32\times 32$ and $64\times 64$, so directly upsampling them to match the image resolution on ImageNet may not be a good way for our method. In addition, STL10 has similar categories as CIFAR10 but a larger image resolution, and thus our method achieves better performance in STL10.
	\begin{table}[t]
		\small
		\begin{center}
			\setlength{\tabcolsep}{0.8mm}{
				\begin{tabular}{l|c|ccc}  
					{Dataset}& Method & 1-label/class & 4-label/class & 10-label/class \cr  
					\shline
					\multirow{3}{*}{DTD}
					& VUP  & 26.74             & 48.20             & 58.68  \cr
					& UF   & 26.45             & \ 48.71               & 58.34   \cr
					& TUP    & \textbf{32.07}    & \textbf{55.17}    & \textbf{65.57}  \cr 
					\hline
					\multirow{3}{*}{Pet37}
					& VUP  &  41.23           & 68.94          & 77.62   \cr
					& UF  & 40.01                  & 67.03                 & 75.56   \cr
					& TUP    & \textbf{57.28}    & \textbf{73.65}    & \textbf{80.48}  \cr
					\hline
					\multirow{3}{*}{STL10}
					& VUP  & 53.86             & 87.35           & 90.85   \cr
					& VF   & 45.22                 & 72.43                 & 79.34   \cr
					& TUP    & \textbf{67.94}    & \textbf{88.07}    & \textbf{91.39}  \cr
			\end{tabular} }
			\captionof{table}{\small \textbf{Vanilla Unsupervised Pretraining(VUP) vs. Unpervised Finetuning(UF) vs. Target-aware Pretraing(TUP)}. The transfer accuracy is evaluated under the oracle setting.}
			\label{tab:TUP_pretrain}
		\end{center}
		\vspace{-3em}
	\end{table}
	
	\subsection{Ablation Study}
	\begin{figure}[!t]
		\centering
		\includegraphics[width=1.0\linewidth]{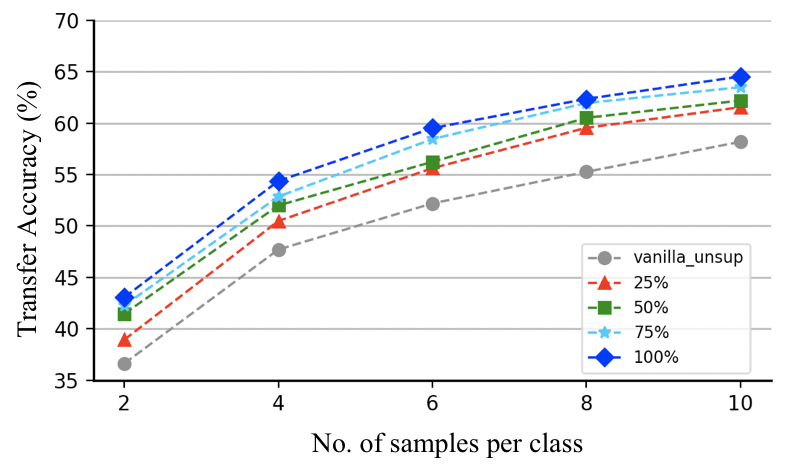}
		\caption{\small \textbf{Different percentages of unlabeled target data} involved into target-aware unsupervised pretraining. The label-efficient transfer performance is used for evaluation.}
		\label{fig:sample_ratio}
		\vspace{-2em}
	\end{figure}
	
	
	\noindent  \textbf{Benefits from Target-aware Unsupervised Pretraining.}
	In this experiment, we validate the advantage of target-aware unsupervised pretraining over vanilla unsupervised pretraining, and another simple unsupervised finetuning baseline. In the unsupervised finetuning baseline, we continue to perform unsupervised pretraining on the unlabeled target set with a smaller learning rate upon the unsupervised pretrained representation on the ImageNet. The results of three representative datasets are shown in the Table \ref{tab:TUP_pretrain}. As we can see, directly unsupervised finetuning can not work well and even degrades the transfer ability of vanilla unsupvised pretraining. Attributing to the merit of simultaneously maintaining the transfer ability learned in large-scale unlabeled source data and involving the information of target set, our target-aware pretraining yields consistent performance gain upon these baselines.
	
	\vspace{0.5em}
	\noindent  \textbf{Influence of Target Dataset Scale in Pretraining.} To further verify the hypothesis that our method will benefits from a larger amount of unlabelled target dataset, we further conduct a simple ablation experiment on the DTD dataset. Specifically, during the target-aware unsupervised pretraining(TUP), we involve different percentages of target data ($25\%, 50\%, 75\%, 100\%$), and then evaluate the few-label transfer performance. As shown in \Fref{fig:sample_ratio}, involving more unlabeled target data into pretraining can help learn better representation, thus producing better performance. 
	
	\noindent \textbf{Ablation of Sample Re-balancing Ratio.} As stated in the method part, the target datasets often have a small image amount and can be smaller than source dataset by several magnitudes. Therefore, we find sample re-balancing is indispensable to relieve the data imbalance issue during pretraining. Here, we use the DTD dataset as an example and try two variants: without sample re-balancing and with a large re-balancing ratio (resampled target dataset size is $50\%$ of the source dataset size). As we can see in \Tref{tab:balance_ratio}, the transfer performance degrades if no sample re-balancing is applied, and too large re-balancing ratio will also lead to inferior results because the benefit from the auxiliary source dataset is suppressed. We empirically find a re-balancing ratio $\sim 20\%$ works all the experiments. 
	\begin{figure}[t]
		\centering
		\includegraphics[width=1.0\linewidth]{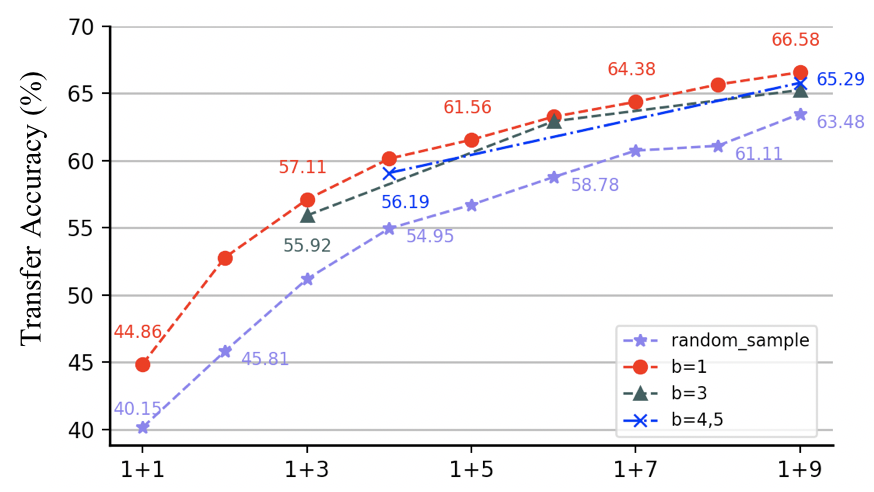}
		\caption{\small \textbf{The number of selected eigen-samples} to query labeling at each evolution step of the progressive few-label transfer. } \label{fig:b_study}
		\vspace{-1em}
	\end{figure}

	\begin{table}[t]
		\small
		\begin{center}
			\setlength{\tabcolsep}{2.3mm}{
				\begin{tabular}{c|ccccc}  
					\shline
					balance ratio & 1+1 & 1+3 &  1+5 & 1+7  &  1+9 \cr  
					\shline 
					w/o re-balance  & 40.27       &  49.55           & 54.42          &  59.28            & 61.01 \cr
					20\%        & \textbf{44.86} &  \textbf{57.11}  & \textbf{61.56} &  \textbf{64.38}   & \textbf{66.58}  \cr
					50\%        & 38.83          &  50.06           & 55.27          &  60.59            & 63.86  \cr 
			\end{tabular} }
			\captionof{table}{\small \textbf{Different sampling re-balancing ratios during pretraining} evaluated by using the performance on the DTD dataset.} \label{tab:balance_ratio}
		\end{center}
		\vspace{-3em}
	\end{table}
	
	\noindent \textbf{Ablation of the Annotation Number $b$.} In our default implementation of progressive few-label transfer, we set $b=1$ at each evolution step. However, we can also set $b>1$ to reduce the total evolution step number and annotate more images at each evolution step. To demonstrate the generalization ability with different $b$ values, we design two simple ablation experiments. In details, suppose the maximum annotation budget is $10 \times C$, we try two different finetuning strategies on the DTD dataset, namely, we either finish the whole progressive process with 3 steps by setting $b=3$ for each step, or with 2 steps by setting $b=4$ for the first step and $b=5$ for the second step. As shown in \Fref{fig:b_study}, these two coarse strategies achieve slightly worse performance than the default finegrained strategy ($b=1$), but still outperform the random sampling baseline by a large margin. By setting $b$ different values, our method can provide the flexibility to achieve a trade-off between performance and training efficiency.
	
	\begin{table}[t]
		\small
		\begin{center}
			\setlength{\tabcolsep}{2.8mm}{
				\begin{tabular}{l|c|cc}  
					{Dataset}& Method & 5-label/class & 10-label/class \cr  
					\shline
					\multirow{2}{*}{CIFAR10}
					& MoCoV2      & 83.64            & 92.15  \cr
					& Ours   & \textbf{87.91}    & \textbf{96.38}  \cr 
					\hline
					\multirow{2}{*}{CIFAR100}
					& MoCoV2     & 61.54            & 70.41  \cr
					& Ours  & \textbf{68.35}    & \textbf{73.77}  \cr
					\hline
					\multirow{2}{*}{Food101}
					& MoCoV2    & 43.26           & 62.98  \cr
					& Ours    &  \textbf{64.83}    & \textbf{75.75}  \cr
					\hline
			\end{tabular} }
			\captionof{table}{\small Semi-supervised Transfer performance comparison between unsupervised pretraining (MoCoV2) and our target-aware unsupervised pretraining.}
			\label{tab:semi-transfer}
		\end{center}
		\vspace{-3em}
	\end{table}
	
	\vspace{0.5em}
	\noindent \textbf{Applying to Semi-supervised Transfer.} In this paper, we mainly focus on improving the few-label transfer performance of unsupervised pretraining and demonstrate the better performance of our target-aware unsupervised pretraining for few-label finetuning. But considering some unlabeled target data is available, we can also leverage some semi-supervised transfer methods to utilize both the unlabeled data and the labeled data. Here we take the SOTA semi-supervised learning method FixMatch \cite{sohn2020fixmatch} and adopt different pretraining models as initialization to compare the final transfer performance. CIFAR10, CIFAR100, and Food101 are used here because they have a larger image number. In this setting, different from our method, the vanilla unsupervised pretraining baseline only leverages the unlabeled data in the transfer stage. As shown in \Tref{tab:semi-transfer}, benefiting from the better pretrained representation as initialization, our method also achieve better semi-supervised transfer performance. 
	
	\section{Conclusion}
	In recent years, unsupervised pretraining has made tremendous progress and many recent works show that unsupervised pretraining can achieve comparable transfer performance to supervised pretraining. But in this paper, we find unsupervised pretraining still perform much worse for few-label transfer, where very few labeled target samples are available for finetuning. This phenomenon has not been studied in existing methods. We provide some possible reasons from the clustering perspective and propose a simple target-aware unsupervised pretraining method to mitigate this issue. This is applicable to the common application scenarios where some small-scale unlabeled data can be collected. To further maximize the few-label transfer performance under a given annotation budget, we also propose a new progressive few-label transfer algorithm, which iteractively finds the best samples to annotate and finetunes the model based on the labeled samples. Through extensive experiments on multiple different datasets, we demonstrate that the proposed strategy can significantly boost the few-label transfer performance of unsupervised pretraining.
	
	\vspace{-1em}
	\section*{Acknowledgement}
	\vspace{-1em}
	This work is supported by the Fundamental Research Funds for the Central Universities (WK2100330002, WK3480000005).
	
	{\small
		\bibliographystyle{ieee_fullname}
		\bibliography{egbib}
	}
	
\end{document}